# Token Manipulation Generative Adversarial Network for Text Generation

DaeJin, Jo
twidddj@gmail.com

## Abstract

MaskGAN opens the query for the conditional language model by filling in the blanks between the given tokens. In this paper, we focus on addressing the limitations caused by having to specify blanks to be filled. We decompose conditional text generation problem into two tasks, make-a-blank and fill-in-the-blank, and extend the former to handle more complex manipulations on the given tokens. We cast these tasks as a hierarchical multi agent RL problem and introduce a conditional adversarial learning that allows the agents to reach a goal, producing realistic texts, in cooperative setting. We show that the proposed model not only addresses the limitations but also provides good results without compromising the performance in terms of quality and diversity.

## 1 Introduction

As a content generator, efficient language model is still required since it is hard to select a useful sample among the huge samples generated. Constraining the sample space to be generated is an effective way to save effort. The constraints are often defined as a set of word tokens, for instance, the headline or summary can be a good constraint for generating a document.

MaskGAN(William et al., 2018) addresses the constraints by training on in-filling task. The seq2-seq (Sutskever et al., 2014) architecture is used to read the masked sequence, providing context to the decoder that auto-regressively fills in missing tokens. When generating conditional samples under the model, the masked tokens to be filled must be specified. This has some limitations as a conditional language model: 1) We must know where and how many the blanks are 2) The model could not change the presented token given appropriately from the previously generated text. In order to overcome these obstacles, we propose a generation process that decides first whether to create a blank and then fills it if it is blank. Otherwise, it utilizes a token given. We present a token manipulation model for this hierarchical cooperative task through adversarial learning.

An experimental evaluation shows that the proposed models provide better performance than MaskGAN on relatively few input tokens even without additional information about what tokens are missing.

The code will be available at https://github.com/twidddj/tokmangan

## 2 MaskGAN

Here we briefly describe about MaskGAN. Pairs of input and output tokens with the same length $L$ are denoted by $\boldsymbol{x} = (x_1, ..., x_L)$ and $\boldsymbol{y} = (y_1, ..., y_L)$ respectively. A mask function $m(x_t)$ generates a binary mask of the same length as $\boldsymbol{m} = (m(x_1), ..., m(x_L))$ where $m(x_t) \in \{0,1\}$, selects which tokens are remain. The generator is defined as

$$G(y_{<t}, \boldsymbol{x}) = \begin{cases} G_\theta(\tilde{y}_t | y_{<t}, \boldsymbol{m}), & if\ m(x_t) = 1 \\ x_t, & otherwise \end{cases}. \quad (1)$$

For an input token $x_t$ that remain, $m(x_t) = 0$, the output token $y_t$ is assigned with $x_t$ and thus it has no contribution on the gradient of both the generator and discriminator at time $t$.

The discriminator has an identical architecture to the generator except that the output is the probability that indicates whether the filled-in token is real or fake. The estimation of the log probability, $\log D(G(\tilde{y}_t = y_t^{real} | y_{<t}, \boldsymbol{m}(\boldsymbol{x})))$, is then regarded as the reinforcement learning reward. The actor-critic architecture is applied to maximize the discounted total return in every generation step.

## 3 TokManGAN

We decompose the conditional probability of generating an output sequence $y$ from an input tokens as follows

$$P(y|x_{SEED}) = \prod_t \sum_{a_t \in \mathcal{A}} P(y_t|y_{<t}, x_{SEED}, a_t) P_{Man}(a_t|y_{<t}, x_{SEED}) \quad (2)$$

where $\mathcal{A}$ is a set of manipulations. Term $x_{SEED}$, which denotes the input tokens with length $L_{SEED}$. We assume that $x_{SEED}$ has the same order of occurrence as $y$. If the manipulations are defined as *'add-a-new-token'* and *'use-a-given-token'*, then it is equivalent to the generator defined as in Equation 1 with a stochastic mask function $P_{Man}(a_t|y_{<t}, x_{SEED})$.

To find the factorized distribution in Equation 2, we consider hierarchical multi agent: 1) *masking manipulator* $G^{Man^+}$ that determines whether to add a blank or not, 2) *token generator* $G^{Tok}$ that infills the blank which occurred by $G^{Man^+}$ and 3) *token manipulator* $G^{Man^-}$ that determines how to utilize current candidate, $x_{SEED}^{idx}$, in given candidates $x_{SEED}$. The output sequence $\{y_i\}_{i=1}^T$ is produced by combining their actions as shown in Figure 1.

A combined $G_\theta$ can be defined according to the manipulations for $G^{Man}$ like as follows:

$$G(y_t) = G_\theta(s_t, a_t) = \begin{cases} \tilde{y}_t \sim G^{Tok}, & \text{if } a_t = 'add' \\ G_\theta(s_t, a_t = 'add'), +idx, & \text{if } a_t = 'replace' \\ x_{SEED}^{idx}, +idx, & \text{if } a_t = 'use' \\ null, +idx, & \text{if } a_t = 'pass' \\ \vdots \end{cases} \quad (3)$$

where $a_t \sim G^{Man}$ and $s_t$ is the environment state that consists of $y_{<t}$, $x_{SEED}$ and $idx$. Term $+idx$ is the operation of increasing $idx$ of $x_{SEED}$.

Our model considers not only for the token generator that fills in missing tokens, but also for the manipulators that select whether it is missing and how the token was used, which is distinct from defining rewards for masked tokens in MaskGAN. The discriminator $D_\phi$ and the reward $r_t$ are defined as respectively,

$$D_\phi(y_t|y_{<t}, x_{SEED}) = P(y_t = y_t^{real}|y_{<t}, x_{SEED}), \quad (4)$$

$$r_t \equiv \log D_\phi(y_t|y_{<t}, x_{SEED}). \quad (5)$$

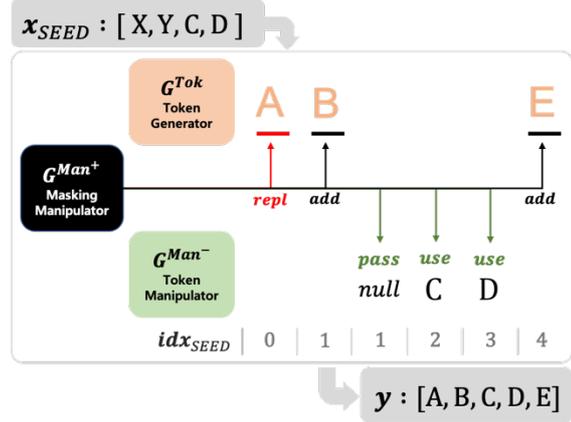

Figure 1: TokManGAN generates target text "A B C D E" from given input tokens "X Y C D".

### 3.1 Token Generator

The token generator $G^{Tok}$ selects the next token $\tilde{y}_t \in \mathcal{Y}$, where $\mathcal{Y}$ is the vocabulary of candidate tokens. We choose the same architecture to the generator in MaskGAN, except that $x_{SEED}$ is not required to have the same length as the output sequence in a masked form. Where the blanks are placed between tokens is determined by the manipulator during in episode. $\tilde{y}_t$ is sampled from a categorical distribution parameterized by $p_t^{Tok} = softmax(f_{linear}^{Tok}(h_t))$ where

$$h_t = attLSTM(y_{<t}, f_{enc}(x_{SEED})). \quad (6)$$

We design $G^{Tok}$ to be rewarded for how realistically the missing tokens are selected for $x_{SEED}$. Therefore, we need information about which part is missing. This is achieved by receiving the real manipulation sequence from the pairs of ($x_{SEED}$, $y$) sampled from the training data and the fake manipulation sequence from the manipulator $G^{Man}$.

### 3.2 Manipulator

The manipulator $G^{Man}$ decomposed hierarchically into two sub-manipulators: 1) *masking manipulator* $G^{Man^+}$ to determine a new token will be added (e.g. $'add'$ or $'replace'$ in Equation 3) or not and 2) *token manipulator* $G^{Man^-}$ to determine how to utilize a given token in $x_{SEED}$ when $G^{Man^+}$ chooses not to add a new token.

To simplify implementation, we treat the sub-manipulators as a combined manipulator so that we can obtain all the manipulations by sampling once. This selects an action $a_t \in \mathcal{A}$ that determines whether to utilize (*use, ignore*, etc.) a token in $x_{SEED}$ or *add* a new token. $a_t$ is sampled from a categorical distribution parameterized by $p_t^{Man} = softmax(f_{linear}^{Man}(h_t))$ where $h_t$ is defined in Equation 6 and $p_t^{Man}$ is a $|\mathcal{A}|$-dimensional probability vector.

The role of this can be seen equivalent to the tagger in *LaserTagger* (Malmi et al., 2019) but the action space of $G^{Man}$ is defined regardless of the size of vocabulary in our setup. This is possible because our model generates tokens simultaneously with manipulations (tagging in *LaserTagger*). Another advantage of this setup is that the manipulator considers $x_{SEED}$ and $y_{<t}$ simultaneously so that our generator could produce more robust results against any form of $x_{SEED}$. For instance, for given two word tokens $x_{SEED} \coloneqq ('the', 'the')$, if the previously generated token $y_{t-1}$ is $'the'$, then it would be unlikely that the current manipulation to be selected as $'use'$.

### 3.3 Training

Here we follow the spirits of language GANs (Yu et al., 2017, William et al., 2018). In our setting, the state value function can be scaled as follows[1]:

$$V^{G_\theta}(s = y_{<t}) = \sum_{y_t \in \mathcal{Y}^+} G_\theta(y_t|y_{<t}) \cdot Q^{G_\theta}$$

$$= \sum_{m_t \in \{0,1\}} G^{Man^+}(m_t|y_{<t}) \sum_{y_t \in \mathcal{A}(m_t)} G^*(y_t|y_{<t}) \cdot Q^{G_\theta}$$

$$= \left[ G^{Man^+}(m_t = 1|y_{<t}) \sum_{\tilde{y}_t \in \mathcal{Y}} G^{Tok}(\tilde{y}_t|y_{<t}) \cdot Q^{G_\theta} \right]$$

$$+ \left[ G_{Man^+}(m_t = 0|y_{<t}) \sum_{a_t \in \mathcal{A}^-} G^{Man^-}(a_t|y_{<t}) \cdot Q^{G_\theta} \right]$$

(7)

where the binary indicator $m_t$ indicates whether the new token at step $t$ is added, $m_t = 1$, or not, $m_t = 0$. The state transition is deterministic after the actions has been chosen, i.e. $\delta_{s,s'}^{m,y} = 1$, for the other next states $s''$, $\delta_{s,s''}^{m,y} = 0$. Then the action value function is defined as follow $Q^{G_\theta}(s = y_{<t}, a = y_t) = \mathcal{R}_s^a + V^{G_\theta}(s') = V^{G_\theta}(y_{1:t})$ as in SeqGAN (Yu et al., 2017).

Following MaskGAN, we apply actor-critic method to reduce the variance of gradient estimator. Our generator seeks to maximize the cumulative total reward $R = \sum_{t=1}^T R_t$ where $R_t = \sum_{s=t}^T \gamma^s r_t$ and $\gamma$ is the discount factor. An unbiased estimator using the learned value function $b_t = V^G(y_{1:t})$ by a critic network is given as $\nabla_\theta \mathbb{E}_G = (R_t - b_t) \nabla_\theta \log G_\theta(y_t)$. By applying the likelihood ratio trick (Williams, 1992), we can obtain the gradients with respect to the parameters of $G^{Tok}$ and $G^{Man2}$:

$$\nabla_T \mathcal{J} = \mathbb{E}_{y_t \sim G^T} \left[ \sum_t A_t \cdot G^+(m_t = 1) \cdot \nabla_T \log G^T(y_t) \right],$$

(8)

$$\nabla_- \mathcal{J} = \mathbb{E}_{a_t \sim G^-} \left[ \sum_t A_t \cdot G^+(m_t = 0) \cdot \nabla_- \log G^-(a_t) \right]$$

(9)

$$\nabla_+ \mathcal{J} = \mathbb{E}_{y_t \sim G^T} \left[ \sum_t A_t \cdot G^T(y_t) \cdot \nabla_+ \log G^+(m_t = 1) \right]$$

$$+ \mathbb{E}_{a_t \sim G^-} \left[ \sum_t A_t \cdot G^-(a_t) \cdot \nabla_+ \log G^+(m_t = 0) \right],$$

(10)

where $A_t = (R_t - b_t)$ is the quantity as an estimate of the advantage function. The gradients intuitively show that the advantage is evaluated against the likelihood of the mutual operations. This evaluator regulates the other connected agent to have more informative rewards and allows that the agents are jointly trained without centralized learning framework such as (Foerster et al., 2018) that may be hard to adapt in extremely large discrete action spaces.

As in conventional GAN training, our discriminator $D_\phi$ is updated by following the gradient

$$\nabla_\phi \frac{1}{m} \left[ \sum_{i=1}^m \log D_\phi(y^{(i)}) + \log\left(1 - D_\phi\left(G(x_{SEED|y^{(i)}})\right)\right) \right].$$

(11)

## 4 Experiments

Here we compare the performance of the proposed models with the models of (William et al., 2018) in different modes (i.e. MLE & GAN). Since the two models have almost the same architectures, the main hyper parameters (such as embedding sizes, unit sizes and number of LSTM layers) that determine the capacity of the model are set identically for evaluation. The conditional samples

---

[1] To minimize notation, we abbreviate the action value function $Q^{G_\theta}(y_{<t}, y_t)$ as $Q^{G_\theta}$ and omit the components of the state; $x_{SEED}$ and $idx$.

[2] The superscript and subscript for the parameters of $G^{Tok}$ are abbreviated as $T$. Likewise, the notations $Man^+$ and $Man^-$ are abbreviated as $+$ and $-$ respectively.

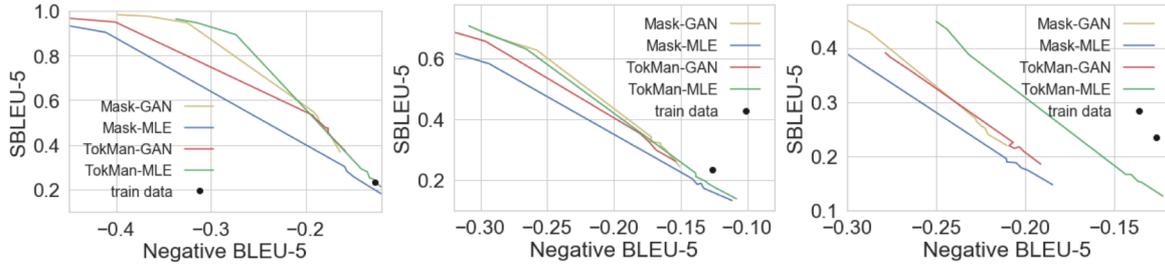

Figure 2: (**Left**) Unconditional results, (**Middle**) Conditional results on *mask rate* = 0.25 and (**Right**) Conditional results on *mask rate* = 0.5. (*lower is better for all metrics*). The scores were calculated for each 6,724 generated samples on test data.

are generated from the input tokens that obtained by changing the mask rate.

### 4.1 Training Setting

**Data:** We use Image COCO dataset (Lin et al., 2014) following the setting (Zhu et al., 2018). We do not use the data with lengths that exceed 20 (≈ 0.01% ). Test sentences including words not included in the training sentences were excluded. After the preprocessing, training set and test set contain 9,895 and 6,724 sentences respectively. The training set has 4,616 distinct words.

**GAN setting:** Both model performs 200 epochs adversarial training after 80 epochs MLE based pre-training. From a given sentence, *mask rate* is chosen randomly between 0.25 to 1. Only the masked information is used as target actions (i.e. *add* and *use*) of the in pre-training.

**Evaluation metrics:** BLEU score (Papineni et al., 2002) and self-BLEU score (Zhu et al., 2018) are used to evaluate quality and diversity respectively. These metrics are compared on quality-diversity space drawn by the temperature sweep (Caccia et al., 2020).

### 4.2 Experiment Results

As we can see in Figure 2, MaskMLE shows better performance than the others on unconditional and conditional setting. Note that as the mask rate is decreased, Mask(MLE&GAN) will produce more realistic results since they already know where the given tokens are presented. TokManGAN shows better performance than MaskGAN on relatively few input tokens (e.g. *mask rate > 0.5*). Considering the fact that our models do not require additional information about the what tokens are missing. Although the proposed model cannot guarantee that a text is generated using all of the given tokens, it can be confirmed that most of the tokens are reflected. If not, the smaller the mask rate, the greater the difference in BLEU score between the two models.

As reported in (Caccia et al., 2020), MaskGAN has scores lower than MaskMLE. However, the results show an interesting fact that TokManGAN has scores better than TokManMLE. This may be a clue that it can negatively affect adversarial learning when the rewards predicted from the discriminator are naively reflected to the generator.

## 5 Conclusions

This paper presented TokManGAN, a hierarchical text generation model with policy gradient under the influence of mutual evaluation. Experimental results show that the proposed model produces good text samples as much as MaskGAN, despite that the model have no additional information about where the given tokens should be placed in the text to be generated.

The manipulations that are not used in pre-training are effectively used in adversarial learning by increasing the dropout rate, but do not appear unless randomness is given in the actual test. This will be investigated in future work.